\documentclass{article} % For LaTeX2e
\usepackage{iclr2019_conference,times}

\usepackage{url}
\usepackage[paperVersion]{marcpaper}

\newcommand{\namingtask}{\textsc{MethodNaming}\xspace}
\newcommand{\doctask}{\textsc{MethodDoc}\xspace}
\newcommand{\nltask}{\textsc{NLSummarization}\xspace}

\newcommand{\bilstmSym}{\textsc{BiLSTM}}
\newcommand{\selfattSym}{\textsc{SelfAtt}}
\newcommand{\bilstm}{\textsc{BiLSTM}\phantom{+GNN}}
\newcommand{\selfatt}{\textsc{SelfAtt}\phantom{+GNN}}
\newcommand{\bilstmGNN}{\bilstmSym+GNN}
\newcommand{\selfattGNN}{\selfattSym+GNN}
\newcommand{\lstm}{\textsc{LSTM}}
\newcommand{\lstmPointer}{\lstm+\textsc{Pointer}}

\newcommand{\seqToseq}[2]{#1~\ensuremath{\rightarrow}~#2}

\title{Structured Neural Summarization}

\author{Patrick Fernandes\thanks{Work done while working at Microsoft Research Cambridge}\\
Carnegie Mellon University \& IT \\
Lisbon, Portugal\\
\texttt{pfernand@cs.cmu.edu} 
\And
Miltiadis Allamanis \& Marc Brockschmidt\\
Microsoft Research\\
Cambridge, United Kingdom\\
\texttt{\{miallama,mabrocks\}@microsoft.com} 
}

\iclrfinalcopy % Uncomment for camera-ready version, but NOT for submission.
\begin{document}

\maketitle

\begin{abstract}
 Summarization of long sequences into a concise statement is a core problem
 in natural language processing, requiring non-trivial understanding of the
 input.
 Based on the promising results of graph neural networks on highly structured
 data, we develop a framework to extend existing sequence encoders with a
 graph component that can reason about long-distance relationships in
 weakly structured data such as text.
 In an extensive evaluation, we show that the resulting hybrid sequence-graph
 models outperform both pure sequence models as well as pure graph models on a
 range of summarization tasks.
\end{abstract}

\section{Introduction}
Summarization, the task of condensing a large and complex input into a
smaller representation that retains the core semantics of the input,
is a classical task for natural language processing systems.
Automatic summarization requires a machine learning component
to identify important entities and relationships between them,
while ignoring redundancies and common concepts.

Current approaches to summarization are based on the
sequence-to-sequence paradigm over the words of some text, with a sequence encoder --- typically
a recurrent neural network, but sometimes a 1D-CNN~\citep{narayan2018ranking} or
using self-attention~\citep{mccann2018natural} --- processing the
input and a sequence decoder generating the output.
Recent successful implementations of this paradigm have substantially
improved performance by focusing on the decoder, extending it with
an attention mechanism over the input sequence and copying
facilities~\citep{see2017get,mccann2018natural}. However,
while standard encoders (\eg bidirectional LSTMs) theoretically
have the ability to handle arbitrary long-distance relationships, in
practice they often fail to correctly handle long texts and are easily
distracted by simple noise~\citep{jia17adversarial}.

In this work, we focus on an improvement of sequence encoders that is
compatible with a wide range of decoder choices.
To mitigate the long-distance relationship problem, we draw inspiration
from recent work on highly-structured objects~\citep{li2015gated,kipf2017semi,gilmer2017neural,allamanis2018learning,cvitkovic2018deep}.
In this line of work, highly-structured data such as entity relationships,
molecules and programs is modelled using graphs.
Graph neural networks are then successfully applied to directly learn
from these graph representations.
Here, we propose to extend this idea to weakly-structured data such as
natural language.
Using existing tools, we can annotate (accepting some noise) such data with additional
relationships (\eg co-references) to obtain a graph.
However, the sequential aspect of the input data is still rich in meaning,
and thus we propose a hybrid model in which a standard sequence encoder
generates rich input for a graph neural network.
In our experiments, the resulting combination outperforms baselines
that use pure sequence or pure graph-based representations.

Briefly, the contributions of our work are:
\begin{enumerate*}
  \item A framework that extends standard sequence encoder models with a
   graph component that leverages additional structure in sequence data.
  \item Application of this extension to a range of existing sequence models
   and an extensive evaluation on three summarization tasks from the literature.
  \item We release all used code and most data at \url{https://github.com/CoderPat/structured-neural-summarization}. The C\# data is not available but is derived from \cite{allamanis2018learning}
\end{enumerate*}

\section{Structured Summarization Tasks}
In this work, we consider three summarization tasks with different properties.
All tasks follow the common pattern of translating a long (structured) sequence into a shorter
sequence while trying to preserve as much meaning as possible.
The first two tasks are related to the summarization of source code (\autoref{fig:codeTasks}),
which is highly structured and thus can profit most from models that can
take advantage of this structure; the final task is a classical natural
language task illustrating that hybrid sequence-graph models are applicable for
less structured inputs as well.

\begin{figure}[t]
    \centering
\begin{subfigure}[b]{\textwidth}
\begin{minipage}{\textwidth}
\begin{lstlisting}[xleftmargin=0cm,basicstyle=\footnotesize\ttfamily,]
public void Add(string name, object value = null, DbType? dbType = null,
                ParameterDirection? direction = null, int? size = null,
                byte? precision = null, byte? scale =null) { 
   parameters[Clean(name)] = new ParamInfo{ 
      Name = name, Value = value, 
      ParameterDirection = direction ?? ParameterDirection.Input, 
      DbType = dbType, Size = size, Precision = precision, Scale = scale
   };}
\end{lstlisting}
\end{minipage}
\end{subfigure}
\\[0.5em]
\begin{subfigure}[b]{\textwidth}
\begin{minipage}{\textwidth} \footnotesize
    \begin{tabular}{ll}
    \textbf{Ground truth}: & add a parameter to this dynamic parameter list\\
    \textbf{\seqToseq{\bilstm}{\lstm}}: &  adds a new parameter to the specified parameter\\
    \textbf{\seqToseq{\bilstmGNN}{\lstm}}: & creates a new instance of the dynamic type specified\\
    \textbf{\seqToseq{\bilstmGNN}{\lstmPointer}}: & add a parameter to a list of parameters
    \end{tabular}
\end{minipage}
\end{subfigure}

    \vspace{-1ex}
    \caption{An example from the dataset for the \doctask source code
    summarization task along with the outputs of a baseline and our models.
    In the \namingtask dataset, this method appears as a sample requiring
    to predict the name \code{Add} as a subtoken sequence of length 1.}\label{fig:codeTasks}
\end{figure}

\paragraph{\namingtask} The aim of this task is to infer the name of a
function (or method in object-oriented languages, such as Java,
  Python and C\#) given its source code~\citep{allamanis2016convolutional}.
Although method names are a single token, they are usually composed of one or
more subtokens (split using \code{snake\_case} or \code{camelCase}) and thus, the
method naming task can be cast as predicting a sequence of subtokens.
Consequently, method names represent an ``extreme'' summary of the
functionality of a given function (on average, the names in the Java dataset
have only 2.9 subtokens).
Notably, the vocabulary of tokens used in names is very large (due to
abbreviations and domain-specific jargon), but this is mitigated by
the fact that 33\% of subtokens in names can be copied
directly from subtokens in the method's source code.
Finally, source code is highly structured input data with known semantics,
which can be exploited to support name prediction.

\paragraph{\doctask}
Similar to the first task, the aim of this task is to predict a succinct
description of the functionality of a method given its source
code~\citep{barone2017parallel}.
Such descriptions usually appear as documentation of methods (\eg 
``docstrings'' in Python or ``JavaDocs'' in Java).
While the task shares many characteristics with the \namingtask task,
the target sequence is substantially longer (on average 19.1
tokens in our C\# dataset) and only 19.4\% of tokens in the
documentation can be copied from the code.
While method documentation is nearer to standard natural language than method names,
it mixes project-specific jargon, code segments and often
describes non-functional aspects of the code, such as performance
characteristics and design considerations.

\paragraph{\nltask}
Finally, we consider the classic summarization of natural language as widely
studied in NLP research.
Specifically, we are interested in abstractive summarization,
where given some text input (\eg a news article) a machine learning model
produces a novel natural language summary.
Traditionally, NLP summarization methods treat text as a sequence of
sentences and each one of them as a sequence of words (tokens).
The input data has less explicitly defined structure than our first two
tasks.
However, we recast the task as a structured summarization problem by
considering additional linguistic structure, including named entities and entity
coreferences as inferred by existing NLP tools.

\section{Model}
As discussed above, standard neural approaches to summarization follow the
sequence-to-sequence framework.
In this setting, most decoders only require a representation $\vect{h}$ of
the complete input sequence (\eg the final state of an RNN) and per-token
representations $\vect{h}_{t_i}$ for each input token $t_i$.
These token representations are then used as the ``memories'' of an attention 
mechanism~\citep{bahdanau2014neural,luong2015effective}
or a pointer network~\citep{vinyals2015pointer}.

In this work, we propose an extension of sequence encoders that allows us to
leverage known (or inferred) relationships among elements in the input data.
To achieve that, we combine sequence encoders with graph neural networks 
(GNNs)~\citep{li2015gated,gilmer2017neural,kipf2017semi}.
For this, we first use a standard sequential encoder (\eg bidirectional RNNs)
to obtain a per-token representation $\vect{h}_{t_i}$, which we then feed
into a GNN as the initial node representations.
The resulting per-node (\ie per-token) representations $\vect{h}'_{t_i}$
can then be used by an unmodified decoder.
Experimentally, we found this to surpass models that use either only the
sequential structure or only the graph structure (see \rSC{sect:eval}).
We now discuss the different parts of our model in detail.

\paragraph{Gated Graph Neural Networks}
To process graphs, we follow \cite{li2015gated} and briefly summarize the
core concepts of GGNNs here.
A graph $\graph = (\nodes, \edgelist, \nodefeats)$ is composed of a set of nodes
$\nodes$, node features $\nodefeats$, and a list of directed edge sets
$\edgelist = (\edges_1, \ldots, \edges_K)$ where $K$ is the number of edge
types.
Each $\node \in \nodes$ is associated with a real-valued vector $\nodefeat_{\node}$
representing the features of the node (\eg, the embedding of a string
label of that node), which is used for the initial state $\state{\node}{0}$
of a node.

Information is propagated through the graph using neural message passing~\citep{gilmer2017neural}.
For this, every node $\node$ sends messages to its neighbors by transforming
its current representation $\state{\node}{i}$ using an edge-type dependent
function $f_{k}$.
Here, $f_{k}$ can be an arbitrary function; we use a simple linear layer.
By computing all messages at the same time, all states can be updated
simultaneously.
In particular, a new state for a node $\node$ is computed by aggregating all
incoming messages as
$\inmsgs{\node}{i} = g(\{ f_k(\state{u}{i}) 
                          \mid \text{there is an edge of type } k \text{ from } u \text{ to } v \})$.
$g$ is an aggregation function; we use elementwise summation for $g$.
Given the aggregated message $\inmsgs{\node}{i}$ and the current state
vector $\state{\node}{i}$ of node $\node$, we can compute the new
state 
$\state{\node}{i+1} = \textsc{GRU}(\inmsgs{\node}{i}, \state{\node}{i})$,
where $\textsc{GRU}$ is the recurrent cell function of a gated recurrent unit.
These dynamics are rolled out for a fixed number of timesteps $T$, and the 
state vectors resulting from the final step are used as output node
representations, \ie, 
$\textsc{GNN}((\nodes, \edgelist, \nodefeats)) = \{ \state{\node}{T} \}_{\node \in \nodes}$.

\paragraph{Sequence GNNs}
We now explain our novel combination of GGNNs and standard sequence encoders.
As input, we take a sequence $S=[s_1 \dots s_N]$ and $K$ binary relationships
$R_1 \ldots R_K \in S \times S$ between elements of the sequence.
For example, $R_=$ could be the equality relationship $\{ (s_i, s_j) \mid s_i = s_j \}$.
The choice and construction of relationships is dataset-dependent, and will be discussed
in detail in \rSC{sect:eval}.
Given any sequence encoder $\mathcal{SE}$ that maps $S$ to per-element representations
$[\mathbf{e}_1 \ldots \mathbf{e}_N]$ and a sequence representation $\mathbf{e}$ (\eg a bidirectional RNN),
we can construct the sequence GNN $\mathcal{SE}_{GNN}$ by simply
computing $[\mathbf{e}_1' \ldots \mathbf{e}_N'] = \textsc{GNN}((S, [R_1 \ldots R_K], [\mathbf{e}_1 \ldots \mathbf{e}_N]))$.
To obtain a graph-level representation, we use the weighted averaging mechanism
from \cite{gilmer2017neural}.
Concretely, for each node $\node$ in the graph, we compute a weight
$\sigma(w(\state{\node}{T})) \in [0,1]$ using a learnable function $w$ and the logistic
sigmoid $\sigma$ and compute a graph-level representation as 
 $\hat{\mathbf{e}} = \sum_{1 \leq i \leq N} \sigma(w(\mathbf{e}_i')) \cdot \aleph(\mathbf{e}_i')$,
where $\aleph$ is another learnable projection function.
We found that best results were achieved by computing the final $\mathbf{e}'$ as
$W \cdot (\mathbf{e}\ \hat{\mathbf{e}})$ for some learnable matrix $W$.

This method can easily be extended to support additional nodes not present
in the original sequence $S$ after running $\mathcal{SE}$ (\eg, to accommodate 
meta-nodes representing sentences, or non-terminal nodes from a syntax tree).
The initial node representation for these additional nodes can come from other
sources, such as a simple embedding of their label.

\paragraph{Implementation Details.}
Processing large graphs of different shapes efficiently requires to overcome
some engineering challenges.
For example, the CNN/DM corpus has (on average) about 900 nodes per graph.
To allow efficient computation, we use the trick of \citet{allamanis2018learning}
where all graphs in a minibatch are ``flattened'' into a single graph with multiple
disconnected components.
The varying graph sizes also represent a problem for the attention and
copying mechanisms in the decoder, as they require to compute a softmax
over a variable-sized list of memories.
To handle this efficiently without padding, we associate each node in
the (flattened) ``batch'' graph with the index of the sample in the minibatch
from which the node originated.
Then, using TensorFlow's \code{unsorted\_segment\_*} operations, we can
perform an efficient and numerically stable softmax over the variable number of
 representations of the nodes of each graph.

\section{Evaluation}

\label{sect:eval}

\subsection{Quantitative Evaluation}
We evaluate Sequence GNNs on our three tasks by comparing them to models that
use only sequence or graph information, as well as by comparing them to
task-specific baselines.
We discuss the three tasks, their respective baselines and how we present the
data to the models (including the relationships considered in the graph
component) next before analyzing the results.

\subsubsection{Setup for \namingtask}
\paragraph{Datasets, Metrics, and Models.}
We consider two datasets for the \namingtask task.
First, we consider the ``Java (small)'' dataset of \citet{alon2018code2seq},
re-using the train-validation-test splits they have picked.
We additionally generated a new dataset from 23 open-source C\# projects
mined from GitHub (see below for the reasons for this second dataset),
removing any duplicates.
More information about these datasets can be found in \autoref{app:datasets}.
We follow earlier work on \namingtask~\citep{allamanis2016convolutional,alon2018code2seq}
and measure performance using the F1 score over the generated subtokens.
However, since the task can be viewed as a form of (extreme) summarization,
we also report ROUGE-2 and ROUGE-L scores~\citep{lin2004rouge}, which we
believe to be additional useful indicators for the quality of results. 
ROUGE-1 is omitted since it is equivalent to F1 score.
We note that there is no widely accepted metric for this task and further
work identifying the most appropriate metric is required.

We compare to the current state of the art~\citep{alon2018code2seq}, as
well as a sequence-to-sequence implementation from the OpenNMT
project~\citep{klein2017opennmt}.
Concretely, we combine two encoders (a bidirectional LSTM encoder with 1
layer and 256 hidden units, and its sequence GNN extension with 128
hidden units unrolled over 8 timesteps)
with two decoders (an LSTM decoder with 1 layer and 256 hidden units
with attention over the input sequence, and an extension using a pointer
network-style copying mechanism~\citep{vinyals2015pointer}).
Additionally, we consider self-attention as an alternative to RNN-based
sequence encoding architectures.
For this, we use the Transformer~\citep{vaswani2017attention} implementation
in OpenNMT (\ie, using self-attention both for the decoder and the encoder)
as a baseline and compare it to a version whose encoder is extended with a
GNN component.

\paragraph{Data Representation}
Following the work of \citet{allamanis2016convolutional,alon2018code2seq},
we break up all identifier tokens (\ie variables, methods, classes, \etc)
in the source code into subtokens by splitting them according to
 \code{camelCase} and \code{pascal\_case} heuristics.
This allows the models to extract information from the information-rich
subtoken structure, and ensures that a copying mechanism in the decoder can
directly copy relevant subtokens, something that we found to be very
effective for this task.
All models are provided with all (sub)tokens belonging to the source code
of a method, including its declaration, with the actual method name
replaced by a placeholder symbol.

To construct a graph from the (sub)tokens, we implement a simplified form
of the work of \citet{allamanis2018learning}.
First, we introduce additional nodes for each (full) identifier token, and
connect the constituent subtokens appearing in the input sequence using a
\textsc{InToken} edge; we additionally connect these nodes using a 
\textsc{NextToken} edge.
We also add nodes for the parse tree and use edges to indicate that one
node is a \textsc{Child} of another.
Finally, we add \textsc{LastLexicalUse} edges to connect identifiers to
their most (lexically) recent use in the source code.
%We also experimented with the full set of edges used by
%\citet{allamanis2018learning}, see \rSC{sect:eval-variations}.

\subsubsection{Setup for \doctask}
\paragraph{Datasets, Metrics, and Models.}
We tried to evaluate on the Python dataset of \citet{barone2017parallel}
that contains pairs of method declarations and their documentation
(``docstring'').
However, following the work of \citet{lopes2017dejavu}, we found extensive
duplication between different folds of the
dataset and were only able to reach comparable results by substantially
overfitting to the training data that overlapped with the test set.
We have documented details in \autoref{app:pythonduplicates} and in \citet{allamanis2018adverse}, and decided
to instead evaluate on our new dataset of 23 open-source C\# projects
from above, again removing duplicates and methods without documentation.
Following \citet{barone2017parallel}, we measure the BLEU score for all models.
However, we also report F1, ROUGE-2 and ROUGE-L scores, which should
better reflect the summarization aspect of the task.
We consider the same models as for the \namingtask task, using the same
configuration, and use the same data representation.

\subsubsection{Setup for \nltask}
\paragraph{Datasets, Metrics, and Models.}
We use the CNN/DM dataset~\citep{hermann2015teaching} using the exact data and split provided
by \citet{see2017get}.
The data is constructed from CNN and Daily Mail news articles along with a few
sentences that summarize each article.
To measure performance, we use the standard ROUGE metrics.
We compare our model with the near-to-state-of-the-art work of \citet{see2017get},
who use a sequence-to-sequence model with attention and copying as basis, but
have additionally substantially improved the decoder component.
As our contribution is entirely on the encoder side and our model uses a
standard sequence decoder, we are \emph{not} expecting to outperform more recent 
models that introduce substantial novelty in the structure or training objective
of the decoder~\citep{chen2018fast,narayan2018ranking}.
Again, we evaluate our contribution using an OpenNMT-based encoder/decoder
combination.
Concretely, we use a bidirectional LSTM encoder with 1 layer and 256
hidden units, and its sequence GNN extension with 128 hidden units unrolled
over 8 timesteps.
As decoder, we use an LSTM with 1 layer and 256 hidden units
with attention over the input sequence, and an extension using a pointer
network-style copying mechanism.

\begin{figure}[t]
    \resizebox{\linewidth}{!}{%
\begin{tikzpicture}[wordgraph]
    \node[token] (s1-t1) at (0,0)
      {Munster};
    \node[token] (s1-t2) at ($(s1-t1.east) + (\tokenhdist, 0)$)
      {have};
    \node[token] (s1-t3) at ($(s1-t2.east) + (\tokenhdist, 0)$)
      {signed};
    \node[token] (s1-t4) at ($(s1-t3.east) + (\tokenhdist, 0)$)
      {New};
    \node[token] (s1-t5) at ($(s1-t4.east) + (\tokenhdist, 0)$)
      {Zealand};
    \node[token] (s1-t6) at ($(s1-t5.east) + (\tokenhdist, 0)$)
      {international};
    \node[token] (s1-t7) at ($(s1-t6.east) + (\tokenhdist, 0)$)
      {Francis};
    \node[token] (s1-t8) at ($(s1-t7.east) + (\tokenhdist, 0)$)
      {Saili};
    \node[token] (s1-t9) at ($(s1-t8.east) + (\tokenhdist, 0)$)
      {on};
    \node[token] (s1-t10) at ($(s1-t9.east) + (\tokenhdist, 0)$)
      {a};
    \node[token] (s1-t11) at ($(s1-t10.east) + (\tokenhdist, 0)$)
      {two-year};
    \node[token] (s1-t12) at ($(s1-t11.east) + (\tokenhdist, 0)$)
      {deal};

    \node[token, anchor=north west] (s2-t1) at ($(s1-t1.south west) + (0, -1.575)$)
      {Utility};
    \node[token] (s2-t2) at ($(s2-t1.east) + (\tokenhdist, 0)$)
      {back};
    \node[token] (s2-t3) at ($(s2-t2.east) + (\tokenhdist, 0)$)
      {Saili};
    \node[token] (s2-t4) at ($(s2-t3.east) + (\tokenhdist, 0)$)
      {who};
    \node[token] (s2-t5) at ($(s2-t4.east) + (\tokenhdist, 0)$)
      {made};
    \node[token] (s2-t6) at ($(s2-t5.east) + (\tokenhdist, 0)$)
      {his};
    \node[] (s2-t7) at ($(s2-t6.east) + (\tokenhdist, 0)$)
      {\ldots};

    \node[entity, anchor=south] (s1) at ($(s1-t6.north) + (0, .5)$)
      {Sentence};
    \node[entity, anchor=north] (s2) at ($(s2-t6.south) + (0, -.5)$)
      {Sentence};
    \draw[->]
      ($(s1.north) + (-.5,0)$)
      -- ($(s1.north west) + (-7, 0)$)
      |- (s2.west);
    \foreach \i in {1,...,12}
    {
        \ifnum\numexpr\i>1
            \pgfmathtruncatemacro{\iOld}{\i - 1};
            \draw (s1-t\iOld.east) edge[nextEdge] (s1-t\i);
        \fi
        \ifnum\numexpr\i<5
            \pgfmathtruncatemacro{\angle}{15-(3*\i)}
            \draw (s1-t\i) edge[inEdge, bend left=\angle] (s1);
        \else
            \ifnum\numexpr\i>8
                \pgfmathtruncatemacro{\angle}{15-(3*(12-\i))}
                \draw (s1-t\i) edge[inEdge, bend right=\angle] (s1);
            \else
                \draw (s1-t\i) edge[inEdge] (s1);
            \fi
        \fi
    }
    \foreach \i in {1,...,7}
    {
        \ifnum\numexpr\i>1
            \pgfmathtruncatemacro{\iOld}{\i - 1};
            \draw (s2-t\iOld.east) edge[nextEdge] (s2-t\i);
        \fi
        \ifnum\numexpr\i<5
            \pgfmathtruncatemacro{\angle}{15-(3*\i)}
            \draw (s2-t\i) edge[inEdge, bend right=\angle] (s2);
        \fi
    }

    \node[entity, anchor=north] (ent-person-1-1) at ($(s1-t1.south) + (0, -.5)$)
      {Person};
    \draw (s1-t1) edge[inEdge] (ent-person-1-1);
    \node[entity, anchor=north] (ent-person-1-7) at ($(s1-t7.south) + (0, -.5)$)
      {Person};
    \draw (s1-t7) edge[inEdge] (ent-person-1-7);
    \node[entity, anchor=south] (ent-person-2-3) at ($(s2-t3.north) + (0, .5)$)
      {Person};
    \draw (s2-t3) edge[inEdge] (ent-person-2-3);
    \node[entity, anchor=north] (ent-country-1-4) at ($(s1-t4.south) + (0, -.5)$)
      {Country};
    \draw (s1-t4) edge[inEdge] (ent-country-1-4);
    \node[entity, anchor=north] (ent-duration-1-11) at ($(s1-t11.south) + (0, -.5)$)
      {Duration};
    \draw (s1-t11) edge[inEdge] (ent-duration-1-11);

    \draw[refEdge]
      (s2-t6)
      -- ($(s2-t6.north) + (0, 0.2)$)
      -| ($(s2-t3.north) + (0.2, 0)$);
    \draw[refEdge]
      ($(s2-t3.north) + (0.1, 0)$)
      -- ($(s2-t3.north) + (0.1, 0.3)$)
      -| ($(s1-t7.west) + (-.2, -.5)$)
      -| ($(s1-t7.south) + (-.2, 0)$);

    \node[token, anchor=north] (ex-token) at ($(ent-person-1-7.south) + (3.2, -.5)$) {Token};
    \node[entity, anchor=north] (ex-entity) at ($(ex-token.south) + (0, -.2)$) {Entity};

    \node[figureLabel, anchor=west, minimum height=0] (label-nextEdge) at ($(ex-token.east) + (1, .1)$) 
      {\textsc{Next}};
    \draw[nextEdge,-] ($(label-nextEdge.west) + (-0.4,0)$) edge[->] ($(label-nextEdge.west) + (-.1, 0)$);
    \node[figureLabel, anchor=north west, minimum height=0] (label-inEdge) at ($(label-nextEdge.south west) + (0, -0.1)$) 
      {\textsc{In}};
    \draw[inEdge,-] ($(label-inEdge.west) + (-0.4,0)$) edge[->] ($(label-inEdge.west) + (-.1, 0)$);    
    \node[figureLabel, anchor=north west, minimum height=0] (label-refEdge) at ($(label-inEdge.south west) + (0, -0.1)$) 
      {\textsc{Ref}};
    \draw[refEdge,-] ($(label-refEdge.west) + (-0.4,0)$) edge[->] ($(label-refEdge.west) + (-.1, 0)$);    

    \begin{pgfonlayer}{bg}
      \draw[dashed, color=black!25, fill=gray!10]
        ($(ex-token.north west) + (-.5, .2)$)
        rectangle
        ($(label-refEdge.south east) + (.3, -.2)$);
    \end{pgfonlayer}
\end{tikzpicture}%
}
    \vspace{-3ex}
    \caption{(Partial) graph of an example input from the CNN/DM corpus.}\label{fig:nlgraph}
\end{figure}

\paragraph{Data Representation}
We use Stanford CoreNLP~\citep{manning2014stanford} (version 3.9.1) to tokenize
the text and provide the resulting tokens to the encoder.
For the graph construction (\autoref{fig:nlgraph}), we extract the named entities and run coreference
resolution using CoreNLP.
We connect tokens using a \textsc{Next} edge and introduce additional
super-nodes for each sentence, connecting each token to the corresponding
sentence-node using a \textsc{In} edge.
We also connect subsequent sentence-nodes using a \textsc{Next} edge.
Then, for each multi-token named entity we create a new node, labeling it with
the type of the entity and connecting it with all tokens referring to that entity
using an \textsc{In} edge.
Finally, coreferences of entities are connected with a special \textsc{Ref} edge.
\autoref{fig:nlgraph} shows a partial graph for an article in the CNN/DM dataset.
The goal of this graph construction process is to explicitly annotate important
relationships that can be useful for summarization.
We note that (a) in early efforts we experimented with adding dependency parse
edges, but found that they do not provide significant benefits and (b) that since
we retrieve the annotations from CoreNLP, they can contain errors and thus, the
performance of the our method is influenced by the accuracy of the upstream
annotators of named entities and coreferences.

\newcommand{\B}[1]{\textbf{#1}}
\begin{table}[t]
\centering
\caption{Evaluation results for all models and tasks.}\label{tbl:results}
\vspace{-1ex}
\footnotesize
\begin{tabular}{@{}llrrrr@{}} \toprule
\multicolumn{2}{@{}l}{\namingtask}   &   F1  & ROUGE-2 & ROUGE-L\\
\midrule

\multicolumn{3}{l}{\underline{Java}} \\
&\citet{alon2018code2seq}            &  43.0  &      -- &   -- \\
&\seqToseq{\selfatt}{\selfatt}       &  24.9  &     8.3 & 27.4 \\
&\seqToseq{\selfattGNN}{\selfatt}    &  44.5  &    20.9 & 43.4 \\
&\seqToseq{\bilstm}{\lstm}           &  35.8  &    17.9 & 39.7 \\
&\seqToseq{\bilstmGNN}{\lstm}        &  44.7  &    21.1 & 43.1 \\
&\seqToseq{\bilstm}{\lstmPointer}    &  42.5  &    22.4 & 45.6 \\
&\seqToseq{\phantom{\bilstmSym+}GNN}{\lstmPointer}  & 50.5 & 24.8 & 48.9 \\
&\seqToseq{\bilstmGNN}{\lstmPointer} &\B{51.4}&\B{25.0} &\B{50.0}\\

\multicolumn{3}{l}{\underline{C\#}} \\
&\seqToseq{\selfatt}{\selfatt}       &   41.3 &   25.2 &   43.2  \\
&\seqToseq{\selfattGNN}{\selfatt}    &   62.1 &   31.0 &   61.1 \\
&\seqToseq{\bilstm}{\lstm}           &   48.8 &\B{32.8}&   51.8  \\
&\seqToseq{\bilstmGNN}{\lstm}        &   62.6 &   31.0 &   61.3  \\
&\seqToseq{\bilstm}{\lstmPointer}    &   57.2 &   29.7 &   60.4  \\
&\seqToseq{\phantom{\bilstmSym+}GNN}{\lstmPointer}  & 63.0 & 31.5 & 61.3\\
&\seqToseq{\bilstmGNN}{\lstmPointer} &\B{63.4}&   31.9 &\B{62.4}\\

% \multicolumn{3}{l}{\underline{C\#}} \\
% &\seqToseq{\bilstm}{\lstm}           &  48.8 &    20.0 & 51.8 \\
% &\seqToseq{\bilstm}{\lstmPointer}    &   ??? &     ??? &  ??? \\
% &\seqToseq{\bilstmGNN}{\lstm} (all edges)       &  64.8 &    33.2 & 63.6 \\
% &\seqToseq{\bilstmGNN}{\lstmPointer} (all edges) &  65.8 &    34.5 & 64.8 \\ 

\midrule
\multicolumn{2}{@{}l}{\doctask}      &     F1  & ROUGE-2 & ROUGE-L & BLEU \\
\midrule

\multicolumn{3}{l}{\underline{C\#}}  \\
&\seqToseq{\selfatt}{\selfatt}       &    40.0 &    27.8 & \B{41.1} & 13.9 \\
&\seqToseq{\selfattGNN}{\selfatt}    &    37.6 &    25.6 &    37.9 & 21.4 \\
&\seqToseq{\bilstm}{\lstm}           &    35.2 &    15.3 &    30.8 & 10.0 \\
&\seqToseq{\bilstmGNN}{\lstm}        &    41.1 &    28.9 &    41.0 & \B{22.5} \\
&\seqToseq{\bilstm}{\lstmPointer}    &    35.2 &    20.8 &   36.7   &  14.7\\
&\seqToseq{\phantom{\bilstmSym+}GNN}{\lstmPointer}  &    38.9 &    25.6 &    37.1 &  17.7 \\
&\seqToseq{\bilstmGNN}{\lstmPointer} (average pooling)  &    43.2 & \B{29.0} &   41.0   & 21.3 \\
&\seqToseq{\bilstmGNN}{\lstmPointer} & \B{45.4} &    28.3 & \B{41.1} &  22.2 \\

\midrule
\multicolumn{2}{@{}l}{\nltask}       & ROUGE-1 & ROUGE-2 & ROUGE-L \\
\midrule

\multicolumn{3}{l}{\underline{CNN/DM}} \\

% &\citet{see2017get}                       &    31.3 &    11.8 &    28.8 \\
&\seqToseq{\bilstm}{\lstm}                &    33.6 &    11.4 &    27.9 \\
&\seqToseq{\bilstmGNN}{\lstm}             &    33.0 &    13.3 &    28.3 \\
&\citet{see2017get} (+ Pointer)           &    36.4 &    15.7 &    33.4 \\
&\seqToseq{\bilstm}{\lstmPointer}         &    35.9 &    13.9 &    30.3 \\
&\seqToseq{\bilstmGNN}{\lstmPointer}      &    38.1 &    16.1 &    33.2 \\
&\citet{see2017get} (+ Pointer + Coverage)&\B{39.5} & \B{17.3}& \B{36.4} \\ 
\bottomrule
\end{tabular}
\vspace{-1ex}
\end{table}

\subsubsection{Results \& Analysis}
We show all results in \rTab{tbl:results}.
Results for models from the literature are taken from the respective papers
and repeated here.
Across all tasks, the results show the advantage of our hybrid sequence GNN
encoders over pure sequence encoders.

On \namingtask, we can see that all GNN-augmented models are able to 
outperform the current specialized state of the art, requiring only simple
graph structure that can easily be obtained using existing parsers for a
programming language.
The results in performance between the different encoder and decoder
configurations nicely show that their effects are largely orthogonal.

On \doctask, the unmodified \seqToseq{\selfattSym}{\selfattSym} model already
performs quite well, and the augmentation with graph data only improves the
BLEU score and worsens the results on ROUGE.
Inspection of the results shows that this is due to the length of predictions.
Whereas the ground truth data has on average 19 tokens in each result,
\seqToseq{\selfattSym}{\selfattSym} predicts on average 11 tokens, and
\seqToseq{\selfattGNN}{\selfattSym} 16 tokens.
Additionally, we experimented with an ablation in which a model is
\emph{only} using graph information, \eg, a setting comparable
to a simplification of the architecture of \citet{allamanis2018learning}.
For this, we configured the GNN to use 128-dimensional representations
and unrolled it for 10 timesteps, keeping the decoder configuration
as for the other models.
The results indicate that this configuration performs less well than
a pure sequenced model.
We speculate that this is mainly due to the fact that 10 timesteps
are insufficient to propagate information across the whole graph, especially
in combination with summation as aggregation function for messages in
graph information propagation.

Finally, on \nltask, our experiments show that the same model suitable
for tasks on highly structured code is competitive with specialized 
models for natural language tasks.
While there is still a gap to the best configuration of \citet{see2017get}
(and an even larger one to more recent work in the area), we believe
that this is entirely due to our simplistic decoder and training objective,
and that our contribution can be combined with these advances.

\begin{table}[t]
    \centering
    \caption{Ablations on CNN/DM Corpus}\label{tbl:ablationNL}
    \vspace{-1ex}
    \footnotesize
    \begin{tabular}{@{}lrrr@{}} \toprule
    
    \nltask (CNN/DM)       & ROUGE-1 & ROUGE-2 & ROUGE-L \\
    \midrule    
    \citet{see2017get} (base)                &    31.3 &    11.8 &    28.8 \\
    \citet{see2017get} (+ Pointer)           &    36.4 &    15.7 &    33.4 \\
    \citet{see2017get} (+ Pointer + Coverage)&\B{39.5} & \B{17.3}& \B{36.4} \\ 
    \midrule
    \seqToseq{\bilstm}{\lstm}                &    33.6 &    11.4 &    27.9 \\
    \seqToseq{\bilstm}{\lstmPointer}         &    35.9 &    13.9 &    30.3 \\
    \seqToseq{\bilstm}{\lstmPointer} (+ coref/entity annotations) &    36.2 &    14.2 &   30.5 \\
    % \seqToseq{\bilstm}{\lstmPointerCov}      &    36.0 &    14.4 &    30.5 \\
    
    \seqToseq{\bilstmGNN}{\lstm}             &    33.0 &    13.3 &    28.3 \\

    \seqToseq{\bilstmGNN}{\lstmPointer} (only sentence nodes)          &    36.0 &    15.2 &    29.6 \\
    \seqToseq{\bilstmGNN}{\lstmPointer} (sentence nodes + eq edges)    &    36.1 &    15.4 &    30.3 \\
    
    \seqToseq{\bilstmGNN}{\lstmPointer}      &    38.1 &    16.1 &    33.2 \\
    % \seqToseq{\bilstmGNN}{\lstmPointerCov}   & \multicolumn{3}{c}{\color{blue}Results pending} \\
    \bottomrule
    \end{tabular}
\vspace{-1ex}
\end{table}

In \autoref{tbl:ablationNL} we show some ablations for \nltask.
As we use the same hyperparameters across all datasets and tasks, 
we additionally perform an experiment with the model of \citet{see2017get}
(as implemented in OpenNMT) but using our settings.
The results achieved by these baselines trend to be a bit worse than the
results reported in the original paper, which we believe is due to a lack
of hyperparameter optimization for this task.
We then evaluated how much the additional linguistic structure provided
by CoreNLP helps.
First, we add the coreference and entity annotations to the baseline
$\textsc{BiLSTM}\rightarrow\lstmPointer$ model (by extending the embedding of
tokens with an embedding of the entity information, and inserting fresh
``<REF1>'', \ldots tokens at the sources/targets of co-references) and observe
only minimal improvements. This suggests that our graph-based encoder is
better-suited to exploit additional structured information compared to a
biLSTM encoder.
We then drop all linguistic structure information
from our model, keeping only the sentence edges/nodes.
This still improves on the baseline $\textsc{BiLSTM}\rightarrow\lstmPointer$
model (in the ROUGE-2 score), suggesting that the GNN
still yields improvements in the absence of linguistic structure. 
Finally, we add long-range dependency
edges by connecting tokens with equivalent string representations of
their stems and observe further minor improvements, indicating that even using
only purely syntactical information, without a semantic parse, can already
provide gains.

\subsection{Qualitative Evaluation}
We look at a few sample suggestions in our dataset across the tasks.
Here we highlight some observations we make that point out interesting
aspects and failure cases of our model.

\begin{figure}[t]
    \centering
\begin{subfigure}[b]{\textwidth}
\begin{minipage}{\textwidth}
\begin{lstlisting}[xleftmargin=0cm,basicstyle=\footnotesize\ttfamily,]
public static bool TryFormat(float value, Span<byte> destination, 
        out int bytesWritten, StandardFormat format=default) {
    return TryFormatFloatingPoint<float>(value, destination,
                                                     out bytesWritten, format); }
\end{lstlisting}
\end{minipage}
\end{subfigure}
\\[0.5em]
\begin{subfigure}[b]{\textwidth}
\begin{minipage}{\textwidth} \footnotesize
    \begin{tabular}{ll}
    \textbf{Ground truth} & formats a single as a utf8 string\\
    \textbf{\seqToseq{\bilstm}{\lstm}} & formats a number of bytes in a utf8 string \\
    \textbf{\seqToseq{\bilstmGNN}{\lstm}} & formats a timespan as a utf8 string \\
    \textbf{\seqToseq{\bilstmGNN}{\lstmPointer}} & formats a float as a utf8 string
    \end{tabular}
\end{minipage}
\end{subfigure}
    \vspace{-1ex}
    \caption{An example from the dataset for the \doctask source code
    summarization task along with the outputs of a baseline and our models.}\label{fig:codeSamples}
\end{figure}

\paragraph{\doctask}
Figures \ref{fig:codeTasks} and \ref{fig:codeSamples} illustrate typical results
of baselines and our model on the \doctask task (see \autoref{app:codesamples}
for more examples).
The hardness of the task stems from the large number of distractors and the need
to identify the most relevant parts of the input.
In \autoref{fig:codeTasks}, the token ``parameter'' and variations appears
many times, and identifying the correct relationship is non-trivial, but is
evidently eased by graph edges explicitly denoting these relationships.
Similarly, in \autoref{fig:codeSamples}, many variables are passed around,
and the semantics of the method require understanding how information flows
between them.

\paragraph{\nltask} \autoref{fig:nlsample} shows one
sample summarization. More samples for this task can be found in \autoref{app:nlsamples}.
First, we notice that the model produces natural-looking summaries with no noticeable negative impact on
the fluency of the language over existing methods.
Furthermore, the GNN-based model seems to capture the central named entity in the article
and creates a summary centered around that entity.
We hypothesize that the GNN component that links long-distance relationships helps capture
and maintain a better ``global'' view of the article, allowing for better identification
of central entities.
Our model still suffers from repetition of information (see \autoref{app:nlsamples}),
and so we believe that our model would also profit from advances such as taking
coverage into account~\citep{see2017get} or optimizing for ROUGE-L scores directly
via reinforcement learning~\citep{chen2018fast,narayan2018ranking}.

\begin{figure}[t]
    \footnotesize
    \begin{tabular}{p{\columnwidth}}\toprule
        \textbf{Input:}
        Arsenal , Newcastle United and Southampton have checked on Caen midfielder N'golo Kante .
        Paris-born Kante is a defensive minded player who has impressed for Caen this season and they are willing to sell for around \pounds~5million .
        Marseille have been in constant contact with Caen over signing the 24-year-old who has similarities with Lassana Diarra and Claude Makelele in terms of stature and style .
        N'Golo Kante is attracting interest from a host of Premier League clubs including Arsenal . Caen would be willing to sell Kante for around \pounds~5million . \\
    \midrule
        \textbf{Reference:}
        n'golo kante is wanted by arsenal , newcastle and southampton .
        marseille are also keen on the \pounds~5m rated midfielder .
        kante has been compared to lassana diarra and claude makelele . click here for the latest premier league news .\\
    \midrule
       \textbf{\citet{see2017get} (+ Pointer):} % 006243_article 
       arsenal , newcastle united and southampton have checked on caen midfielder n'golo kante .
       paris-born kante is attracting interest from a host of premier league clubs including arsenal .
       paris-born kante is attracting interest from a host of premier league clubs including arsenal \\
    \midrule
       \textbf{\citet{see2017get} (+ Pointer + Coverage):} 
       arsenal , newcastle united and southampton have checked on caen midfielder n'golo kante .
       paris-born kante is a defensive minded player who has impressed for caen this season .
       marseille have been in constant contact with caen over signing the 24-year-old .
       \\
    \midrule
        \textbf{\seqToseq{\bilstmGNN}{\lstm}:}
        marseille have been linked with caen midfielder \%UNK\% \%UNK\% .
        marseille have been interested from a host of premier league clubs including arsenal .
        caen have been interested from a host of premier league clubs including arsenal .\\
    \midrule
        \textbf{\seqToseq{\bilstmGNN}{\lstmPointer}}
        n'golo kante is attracting interest from a host of premier league clubs .
        marseille have been in constant contact with caen over signing the 24-year-old .
        the 24-year-old has similarities with lassana diarra and claude makelele in terms of stature . \\ 
        \bottomrule
    \end{tabular}
    \caption{Sample natural language translations from the CNN-DM dataset.}\label{fig:nlsample} 
\end{figure}

\section{Related Work}
Natural language processing research has studied summarization for a
long time.
Most related is work on abstractive summarization, in which the core
content of a given text (usually a news article) is summarized in a
novel and concise sentence.
\citet{chopra2016abstractive} and \citet{nallapati2016abstractive} use
deep learning models with attention on the input text to guide a decoder
that generates a summary.
\citet{see2017get} and \citet{mccann2018natural} extend this idea with
pointer networks~\citep{vinyals2015pointer} to allow for copying tokens
from the input text to the output summary.
These approaches treat text as a simple token sequences, not explicitly
exposing additional structure.
In principle, deep sequence networks are known to be able to learn the
inherent structure of natural language (\eg in
parsing~\citep{vinyals2015grammar} and entity
recognition~\citep{lample2016neural}), but our experiments indicate
that explicitly exposing this structure by separating concerns improves
performance.

Recent work in summarization has proposed improved training objectives
for summarization, such as tracking coverage of the input
document~\citep{see2017get} or using reinforcement learning to directly
identify actions in the decoder that improve target measures such as
ROUGE-L~\citep{chen2018fast,narayan2018ranking}.
These objectives are orthogonal to the graph-augmented encoder discussed
in this work, and we are interested in combining these efforts in future
work.

Exposing more language structure explicitly has been studied over the
last years, with a focus on tree-based models~\citep{tai2015improved}.
Very recently, first uses of graphs in natural language processing have
been explored.
\citet{marcheggiani2017encoding} use graph convolutional networks to
encode single sentences and assist machine translation.
\citet{de2018question} create a graph over named entities over a set of
documents to assist question answering.
Closer to our work is the work of \citet{liu2018toward}, who use abstract
meaning representation (AMR), in which the source document is first parsed
into AMR graphs, before a summary graph is created, which is finally rendered
in natural language.
In contrast to that work we do not use AMRs but directly encode relatively
simple relationships directly on the tokenized text, and do not treat
summarization as a graph rewrite problem.
Combining our encoder with AMRs to use richer graph structures may be a
promising future direction.

Finally, summarization in source code has also been studied in the forms of
method naming, comment and documentation prediction.
Method naming has been tackled with a series of models.
For example, \citet{allamanis2015suggesting} use a log-bilinear network to
predict method names from features,
and later extend this idea to use a convolutional attention network
over the tokens of a method to predict the subtokens of names~\citep{allamanis2016convolutional}.
\citet{raychev2015predicting} and \citet{bichsel2016statistical} use CRFs
for a range of tasks on source code, including the inference of names for
variables and methods.
Recently, \citet{alon2018general,alon2018code2seq} extract and encode
paths from the syntax tree of a program, setting the state of the art
in accuracy on method naming.

Linking text to code can have useful applications, such as code
search~\citep{gu2018deep},
traceability \citep{guo2017semantically}, 
and detection of redundant method comments \citep{louis2018deep}.
Most approaches on source code either treat it as natural language (\ie,
a token sequence), or use a language parser to explicitly expose its
tree structure.
For example, \citet{barone2017parallel} use a simple sequence-to-sequence
baseline, whereas \citet{hu2017codesum} summarize source code by linearizing
the abstract syntax tree of the code and using a sequence-to-sequence
model.
\citet{wan2018improving} instead directly operate on the tree structure
using tree recurrent neural networks~\citep{tai2015improved}.
The use of additional structure on related tasks on source code has
been studied recently, for example in models that are conditioned on
learned traversals of the syntax tree~\citep{bielik2016phog} and
in graph-based approaches~\citep{allamanis2018learning,cvitkovic2018deep}.
However, as noted by \citet{liao2018graph}, GNN-based approaches
suffer from a tension between the ability to propagate information across
large distances in a graph and the computational expense of the propagation
function, which is linear in the number of graph edges per propagation
step.

\section{Discussion \& Conclusions}
We presented a framework for extending sequence encoders with a graph
component that can leverage rich additional structure.
In an evaluation on three different summarization tasks, we have shown
that this augmentation improves the performance of a range of different
sequence models across all tasks.
We are excited about this initial progress and look forward to deeper
integration of mixed sequence-graph modeling in a wide range of tasks
across both formal and natural languages.
The key insight, which we believe to be widely applicable, is that 
inductive biases induced by explicit relationship modeling are a
simple way to boost the practical performance of existing deep
learning systems.


\newpage
\bibliography{bibliography}
\bibliographystyle{iclr2019_conference}

\newpage
\appendix
\section{Code Summarization Samples}
\label{app:codesamples}
\subsection{\doctask}
\textbf{\underline{C\# Sample 1}}
\begin{lstlisting}[xleftmargin=0cm,basicstyle=\footnotesize\ttfamily]
public static bool TryConvertTo(object valueToConvert, Type resultType, 
                                IFormatProvider formatProvider, out object result){
    result = null; 
    try{ 
        result = ConvertTo(valueToConvert, resultType, formatProvider); 
    } catch (InvalidCastException){ 
        return false; 
    } catch (ArgumentException){
        return false; 
    } 
    return true; 
}
\end{lstlisting}
{\footnotesize
\begin{tabular}{ll}
    \textbf{Ground truth} & sets result to valuetoconvert converted to resulttype considering\\
                          & formatprovider for custom conversions calling the parse method \\
                          & and calling convert . changetype .\\
    \textbf{\seqToseq{\bilstm}{\lstm}} & converts the specified type to a primitive type . \\
    \textbf{\seqToseq{\bilstmGNN}{\lstm}} & sets result to resulttype \\
    \textbf{\seqToseq{\bilstmGNN}{\lstmPointer}} & sets result to valuetoconvert converted to resulttype.
\end{tabular}
}

\textbf{\underline{C\# Sample 2}}

\begin{lstlisting}[xleftmargin=0cm,basicstyle=\footnotesize\ttfamily]
public virtual Task Init(string name, IProviderRuntime providerRuntime,
                         IProviderConfiguration config){
    Log = providerRuntime.GetLogger(this.GetType().FullName);
    this.serializerSettings = OrleansJsonSerializer.GetDefaultSerializerSettings(); 
    return TaskDone.Done;
}
\end{lstlisting}
{\footnotesize
\begin{tabular}{ll}
    \textbf{Ground truth} & initializes the storage provider\\
    \textbf{\seqToseq{\bilstm}{\lstm}} & creates a grain object\\
    \textbf{\seqToseq{\bilstmGNN}{\lstm}} & initializes the provider provider\\
    \textbf{\seqToseq{\bilstmGNN}{\lstmPointer}} & initialization function to initialize the specified provider.
\end{tabular}
}

\textbf{\underline{C\# Sample 3}}
\begin{lstlisting}[xleftmargin=0cm,basicstyle=\footnotesize\ttfamily]
public void NullParameter(){
    TaskParameter t = new TaskParameter(null); 
    Assert.Null(t.WrappedParameter); 
    Assert.Equal( TaskParameterType.Null , t.ParameterType ); 
    ((INodePacketTranslatable) t).Translate(
                                TranslationHelpers.GetWriteTranslator());
    TaskParameter t2 = TaskParameter.FactoryForDeserialization(
                                TranslationHelpers.GetReadTranslator()); 
    Assert.Null(t2.WrappedParameter); 
    Assert.Equal(TaskParameterType.Null, t2.ParameterType);
}
\end{lstlisting}
{\footnotesize
\begin{tabular}{ll}
    \textbf{Ground truth} & verifies that construction and serialization with a null parameter is ok\\
    \textbf{\seqToseq{\bilstm}{\lstm}} & tests that the value is a value that is a value to the specified type\\
    \textbf{\seqToseq{\bilstmGNN}{\lstm}} & verifies that construction with an parameter parameter\\
    \textbf{\seqToseq{\bilstmGNN}{\lstmPointer}} & verifies that construction and serialization with a parameter that is null 
\end{tabular}
}

\textbf{\underline{C\# Sample 4}}
\begin{lstlisting}[xleftmargin=0cm,basicstyle=\footnotesize\ttfamily]
public override DbGeometryWellKnownValue CreateWellKnownValue(DbGeometry geometryValue){
  geometryValue.CheckNull("geometryValue"); 
  var spatialValue = geometryValue.AsSpatialValue(); 
  DbGeometryWellKnownValue result = CreateWellKnownValue(spatialValue, 
    ()=>SpatialExceptions.CouldNotCreateWellKnownGeometryValueNoSrid("geometryValue"), 
    ()=>SpatialExceptions.CouldNotCreateWellKnownGeometryValueNoWkbOrWkt("geometryValue"), 
    (srid, wkb, wkt) => new DbGeometryWellKnownValue() {
       CoordinateSystemId = srid, WellKnownBinary = wkb, WellKnownText = wkt 
    }); 
  return result; 
}
\end{lstlisting}
{\footnotesize
\begin{tabular}{ll}
    \textbf{Ground truth}
      & creates an instance of t:system.data.spatial.dbgeometry value using \\
      & one or both of the standard well known spatial formats.\\
    \textbf{\seqToseq{\bilstm}{\lstm}} 
      & creates a t:system.data.spatial.dbgeography value based on the \\
      & specified well known binary value .\\
    \textbf{\seqToseq{\bilstmGNN}{\lstm}}
      & creates a new t:system.data.spatial.dbgeography instance using the \\
      & specified well known spatial formats .\\
    \textbf{\seqToseq{\bilstmGNN}{\lstmPointer}} 
      & creates a new instance of the t:system.data.spatial.dbgeometry value \\
      & based on the provided geometry value and returns the resulting well \\
      & as known spatial formats .\\
\end{tabular}
}

\subsection{\namingtask}
\textbf{\underline{C\# Sample 1}}
\begin{lstlisting}[xleftmargin=0cm,basicstyle=\footnotesize\ttfamily]
public bool _(D d) { 
    return d != null && d.Val == Val ; 
}
\end{lstlisting}
{\footnotesize
\begin{tabular}{ll}
    \textbf{Ground truth}
      & \code{equals}\\
    \textbf{\seqToseq{\bilstm}{\lstm}} 
      & \code{foo}\\
    \textbf{\seqToseq{\bilstmGNN}{\lstm}}
      & \code{equals}\\
    \textbf{\seqToseq{\bilstmGNN}{\lstmPointer}} 
      & \code{equals}\\
\end{tabular}
}

\textbf{\underline{C\# Sample 2}}
\begin{lstlisting}[xleftmargin=0cm,basicstyle=\footnotesize\ttfamily]
internal void _(string switchName, Hashtable bag, string parameterName) {
    object obj = bag[parameterName]; 
    if(obj != null){
        int value = (int) obj; 
        AppendSwitchIfNotNull(switchName, 
                              value.ToString(CultureInfo.InvariantCulture));
    }
}
\end{lstlisting}
{\footnotesize
\begin{tabular}{ll}
    \textbf{Ground truth}
      & \code{append switch with integer}\\
    \textbf{\seqToseq{\bilstm}{\lstm}} 
      & \code{set string}\\
    \textbf{\seqToseq{\bilstmGNN}{\lstm}}
      & \code{append switch}\\
    \textbf{\seqToseq{\bilstmGNN}{\lstmPointer}} 
      & \code{append switch if not null}\\
\end{tabular}
}

\textbf{\underline{C\# Sample 3}}
\begin{lstlisting}[xleftmargin=0cm,basicstyle=\footnotesize\ttfamily]
    internal static string _(){ 
        var currentPlatformString = string.Empty; 
        if (RuntimeInformation.IsOSPlatform(OSPlatform.Windows)){ 
            currentPlatformString = "WINDOWS"; 
        } 
        else if (RuntimeInformation.IsOSPlatform(OSPlatform.Linux)){ 
            currentPlatformString = "LINUX"; 
        } 
        else if ( RuntimeInformation.IsOSPlatform(OSPlatform.OSX)) { 
            currentPlatformString = "OSX"; 
        } 
        else { 
            Assert.True(false, "unrecognized current platform"); 
        } 
        return currentPlatformString ; 
}
\end{lstlisting}
{\footnotesize
\begin{tabular}{ll}
    \textbf{Ground truth}
      & \code{get os platform as string}\\
    \textbf{\seqToseq{\bilstm}{\lstm}} 
      & \code{get name}\\
    \textbf{\seqToseq{\bilstmGNN}{\lstm}}
      & \code{get platform}\\
    \textbf{\seqToseq{\bilstmGNN}{\lstmPointer}} 
      & \code{get current platform string}\\
\end{tabular}
}

\textbf{\underline{C\# Sample 4}}
\begin{lstlisting}[xleftmargin=0cm,basicstyle=\footnotesize\ttfamily]
public override DbGeometryWellKnownValue CreateWellKnownValue(DbGeometry geometryValue){
  geometryValue.CheckNull("geometryValue"); 
  var spatialValue = geometryValue.AsSpatialValue(); 
  DbGeometryWellKnownValue result = CreateWellKnownValue(spatialValue, 
    ()=>SpatialExceptions.CouldNotCreateWellKnownGeometryValueNoSrid("geometryValue"), 
    ()=>SpatialExceptions.CouldNotCreateWellKnownGeometryValueNoWkbOrWkt("geometryValue"), 
    (srid, wkb, wkt) => new DbGeometryWellKnownValue () {
        CoordinateSystemId = srid , WellKnownBinary = wkb , WellKnownText = wkt 
    }); 
  return result; 
}
\end{lstlisting}
{\footnotesize
\begin{tabular}{ll}
    \textbf{Ground truth}
      & \code{create well known value}\\
    \textbf{\seqToseq{\bilstm}{\lstm}} 
      & \code{spatial geometry from xml}\\
    \textbf{\seqToseq{\bilstmGNN}{\lstm}}
      & \code{geometry point}\\
    \textbf{\seqToseq{\bilstmGNN}{\lstmPointer}} 
      & \code{get well known value}\\
\end{tabular}
}

\textbf{\underline{Java Sample 1}}
\begin{lstlisting}[xleftmargin=0cm,basicstyle=\footnotesize\ttfamily]
public static void _(String name, int expected, MetricsRecordBuilder rb) { 
    Assert.assertEquals("Bad value for metric " + name, 
                        expected, 
                        getIntCounter(name, rb));
} 
\end{lstlisting}
{\footnotesize
\begin{tabular}{ll}
    \textbf{Ground truth}
      & \code{assert counter}\\
    \textbf{\seqToseq{\bilstm}{\lstm}} 
      & \code{assert email value}\\
    \textbf{\seqToseq{\bilstmGNN}{\lstm}}
      & \code{assert header}\\
    \textbf{\seqToseq{\bilstmGNN}{\lstmPointer}} 
      & \code{assert int counter}
\end{tabular}
}

\newpage
\section{Natural Language Summarization Samples}
\label{app:nlsamples}
    \begin{tabular}{p{\columnwidth}}\toprule
        \textbf{Input:}
        -LRB- CNN -RRB- Gunshots were fired at rapper Lil Wayne 's tour bus early Sunday in Atlanta .
        No one was injured in the shooting , and no arrests have been made , Atlanta Police spokeswoman Elizabeth Espy said .
        Police are still looking for suspects . Officers were called to a parking lot in Atlanta 's Buckhead neighborhood , Espy said .
        They arrived at 3:25 a.m. and located two tour buses that had been shot multiple times .
        The drivers of the buses said the incident occurred on Interstate 285 near Interstate 75 , Espy said .
        Witnesses provided a limited description of the two vehicles suspected to be involved : a `` Corvette style vehicle '' and an SUV .
        Lil Wayne was in Atlanta for a performance at Compound nightclub Saturday night . CNN 's Carma Hassan contributed to this report .\\
    \midrule
        \textbf{Reference:}
        rapper lil wayne not injured after shots fired at his tour bus on an atlanta interstate , police say .
        no one has been arrested in the shooting \\
    \midrule
    \textbf{\citet{see2017get} (+ Pointer):} % 000758_article
    police are still looking for suspects .
    the incident occurred on interstate 285 near interstate 75 , police say .
    witnesses provided a limited description of the two vehicles suspected to be involved : a `` corvette style vehicle '' and an suv .\\ 
    \midrule
    \textbf{\citet{see2017get} (+ Pointer + Coverage):} 
    lil wayne 's tour bus was shot multiple times , police say .
    police are still looking for suspects .
    they arrived at 3:25 a.m. and located two tour buses that had been shot .\\
    \midrule
        \textbf{\seqToseq{\bilstmGNN}{\lstm}:}
        the incident occurred on interstate \%UNK\% near interstate 75 .
         no one was injured in the shooting , and no arrests have been made , atlanta police spokeswoman says .\\
    \midrule
        \textbf{\seqToseq{\bilstmGNN}{\lstmPointer}}
        gunshots fired at rapper lil wayne 's tour bus early sunday in atlanta , police say .        
        no one was injured in the shooting , and no arrests have been made , police say .\\ \bottomrule
    \end{tabular}

    \vspace{2em}

    \begin{tabular}{p{\columnwidth}}\toprule
        \textbf{Input:}
        Tottenham have held further discussions with Marseille over a potential deal for midfielder Florian Thauvin .
        The 22-year-old has been left out of the squad for this weekend 's game with Metz as Marseille push for a \pounds~15m sale .
        The winger , who can also play behind the striker , was the subject of enquiries from Spurs earlier in the year and has also been watched by Chelsea and Valencia .
        Tottenham have held further talks with Ligue 1 side Marseille over a possible deal for Florian Thauvin .
        Marseille are already resigned to losing Andre Ayew and Andre-Pierre Gignac with English sides keen on both .
        Everton , Newcastle and Swansea , have all shown an interest in Ayew , who is a free agent in the summer .\\
    \midrule
        \textbf{Reference:} 
        florian thauvin has been left out of marseille 's squad with metz .
        marseille are pushing for a \pounds~15m sale and tottenham are interested . the winger has also been watched by chelsea and la liga side valencia .\\
    \midrule
    \textbf{\citet{see2017get} (+ Pointer):} % 005197_article
    tottenham have held further discussions with marseille over a potential deal for midfielder florian thauvin .
    the 22-year-old has been left out of the squad for this weekend 's game with metz as marseille push for a £15m sale .\\
    \midrule
    \textbf{\citet{see2017get} (+ Pointer + Coverage):} 
    florian thauvin has been left out of the squad for this weekend 's game with metz as marseille push for a £15m sale .
    the 22-year-old was the subject of enquiries from spurs earlier in the year .\\
    \midrule
        \textbf{\seqToseq{\bilstmGNN}{\lstm}:}
        the 22-year-old has been left out of the squad for this weekend 's game with metz .
        the 22-year-old has been left out of the squad for this weekend 's game with metz .
        the winger has been left out of the squad for this weekend 's game with metz .
        \\
    \midrule
        \textbf{\seqToseq{\bilstmGNN}{\lstmPointer}}
        tottenham have held further discussions with marseille over a potential deal .
        the winger has been left out of the squad for this weekend 's game .
        tottenham have held further talks with marseille over a potential deal .
        \\ \bottomrule
    \end{tabular}

\newpage
\section{Code Datasets Information}
\label{app:datasets}
\subsection{C\# Dataset}
We extract the C\# dataset from open-source projects on GitHub. Overall,
our dataset contains 460,905 methods, 55,635 of which have a documentation
comment. The dataset is split 85-5-10\%. The projects and \emph{exact} state
of the repositories used is listed in \autoref{tbl:csharpdataset}

\begin{table}[hp]
    \caption{Projects in our C\# dataset. Ordered alphabetically.}\label{tbl:csharpdataset}
    \resizebox{\textwidth}{!}{
    \begin{tabular}{lrp{7cm}} \toprule
    Name & Git SHA & Description \\ \midrule
    Akka.NET        & \code{6f32f6a7} & Actor-based Concurrent \& Distributed Framework\\
    AutoMapper      & \code{19d6f7fc} & Object-to-Object Mapping Library \\
    BenchmarkDotNet & \code{57005f05} & Benchmarking Library\\
    CommonMark.NET  & \code{f3d54530} & Markdown Parser\\
    CoreCLR         & \code{cc5dcbe6} & .NET Core Runtime\\
    CoreFx          & \code{ec1671fd} & .NET Foundational Libraries\\
    Dapper          & \code{3c7cde28} & Object Mapper Library\\
    EntityFramework & \code{c4d9a269} & Object-Relational Mapper\\
    Humanizer       & \code{2b1c94c4} & String Manipulation and Formatting\\
    Lean            & \code{90ee6aae} & Algorithmic Trading Engine\\
    Mono            & \code{9b9e4f4b} & .NET Implementation\\
    MsBuild         & \code{7f95dc15} & Build Engine\\
    Nancy           & \code{de458a9b} & HTTP Service Framework \\
    NLog            & \code{49fdd08e} & Logging Library\\
    Opserver        & \code{9e4d3a40} & Monitoring System\\
    orleans         & \code{f89c5866} & Distributed Virtual Actor Model \\
    Polly           & \code{f3d2973d} & Resilience \& Transient Fault Handling Library \\
    Powershell      & \code{9ac701db} & Command-line Shell \\
    ravendb         & \code{6437de30} & Document Database \\
    roslyn          & \code{8ca0a542} & Compiler \& Code Analysis \& Compilation\\
    ServiceStack    & \code{17f081b9} & Real-time web library\\
    SignalR         & \code{9b05bcb0} & Push Notification Framework\\
    Wox             & \code{13e6c5ee} & Application Launcher\\ \bottomrule
    \end{tabular}
    }
\end{table}

\subsection{Java Method Naming Datasets}
We use the datasets and splits of \citet{alon2018code2seq} provided by their website.
Upon scanning all methods in the dataset, the size of the corpora can be seen
in \autoref{tbl:javadataset}. More information can be found at \citet{alon2018code2seq}.

\begin{table}
    \centering
    \caption{The statistics of the extracted graphs from the Java method naming dataset
    of \citet{alon2018code2seq}.}\label{tbl:javadataset}
    \begin{tabular}{lrrr} \toprule
    Dataset & Train Size & Valid Size & Test Size \\ \midrule
    Java -- Small & 691,505 & 23,837 & 56,952 \\
    %Java -- Large & 16,108,083 & 322,227 & 423,941 \\ 
    \bottomrule
    \end{tabular}
\end{table}

\subsection{Python Method Documentation Dataset}
\label{app:pythonduplicates}
We use the dataset as split of \citet{barone2017parallel} provided by their GitHub repository.
Upon parsing the dataset, we get 106,065 training samples, 1,943 validation samples and 1,937
test samples. We note that 16.9\% of the documentation samples in the validation set and 15.3\% of the
samples in test set have a sample with the identical natural language documentation on the training
set. This eludes to a potential issue, described by \citet{lopes2017dejavu}. See \citet{allamanis2018adverse}
for a lengthier discussion of this issue.

\subsection{Graph Data Statistics}
Below we present the data characteristics of the graphs we use
across the datasets.
\begin{table}[hp]
    \centering
    \caption{Graph Statistics For Datasets.}\label{tbl:graphstats}
    \begin{tabular}{lrr} \toprule
    Dataset & Avg Num Nodes & Avg Num Edges \\ \midrule
    CNN/DM                  & 903.2 & 2532.9\\
   %Gigaword                & 519.6 & 1359.9\\ 
    C\# Method Names        & 125.2 & 239.3\\
    C\# Documentation       & 133.5 & 265.9\\
    Java-Small Method Names & 144.4 & 251.6\\
    %Java-Large Method Names & 127.9 & 205.4\\
    %Python Documentation    & 215.4 & 434.8\\
    \bottomrule
    \end{tabular}
\end{table}

\end{document}